# Image Registration Based Flicker Solving in Video Face Replacement and Analysis Based Sub-pixel Image Registration

Xiaofang Wang [1], Guoqiang Xiang [1*], Xinyue Zhang [1], Wei Wei [1]

[1] School of Computer Science and Engineering, Hebei University of Technology, Beichen, Tianjin 300401, People's Republic of China
[*] xiangguoqiang17@163.com

**Abstract:** In this paper, a framework of video face replacement is proposed and it deals with the flicker of swapped face in video sequence. This framework contains two main innovations: 1) the technique of image registration is exploited to align the source and target video faces for eliminating the flicker or jitter of the segmented video face sequence; 2) a fast sub-pixel image registration method is proposed for farther accuracy and efficiency. Unlike the priori works, it minimizes the overlapping region and takes spatiotemporal coherence into account. Flicker in resulted videos is usually caused by the frequently changed bound of the blending target face and unregistered faces between and along video sequences. The sub-pixel image registration method is proposed to solve the flicker problem. During the alignment process, integer pixel registration is formulated by maximizing the similarity of images with down sampling strategy speeding up the process and sub-pixel image registration is a single-step image match via analytic method. Experimental results show the proposed algorithm reduces the computation time and gets a high accuracy when conducting experiments on different data sets.

## 1. Introduction

Video face replacement is an interesting but challenging task and has been paid much attention by video makers and film producers. As a new research field face replacement technology has been the major topic of researchers and has extensive application value in daily life. Bitouk [1] et al. present a system that automatically swaps faces in images with a large face dataset needed to match the target face in pose and appearance. The technique does not require complex 3D model or any manual intervention. While the face skin or light conditions are different, the result of face swap is natural. Cheng [2] et al. describes an automatic face replacement in video based on 3D morph-able model, which requires a large library of 3D scans of faces. Each face model data consist of 65536 vertices with textures, which would cause a large time and space cost. Feng [3] et al. present a 2D morph-able model based automatic face replacement approach and they adopt active shape models(ASM) for face alignment and face morph for matching source face with the target. Kevin Dale [4] et al. proposed a method to track the facial expression data from the original video clips and target video sequences using multilinear face model, and a little manual intervention is needed. Afifi[5] et al. suggested a system to replace video faces automatically by using the MPB approach to eliminate the bleeding problem and get better replaced face, in which the face alignment process needs users' intervention.

The technique of face replacement in videos is becoming more refined; at the same time the flicker of swapped video faces has been encountered by almost every algorithm and makes the processed videos look unnatural and unsmooth. It may be caused by swapping unregistered faces [1] or replacing faces without time coherency [4]. And it is necessary to align faces between target and source frames and along the frame sequences separately.

Image registration technology is a fundamental and challenging task and is adopted in many applications related to computer vision and image processing [6], such as image mosaic [7], super-resolution reconstruction [8], medical image measurement [9], quantitative target detection and description [10], etc. It makes two or more images have the same features (colour, texture, geometry, or semantic information, etc.) captured at different times and/or using different devices [10].

Given a pair of images (the reference image and unregistered one, respectively) or a video footage, image registration is to find an optimal motion of the latter image (or frame) and make it to have the best similarity with the former. Image registration techniques can be categorized into two parts, feature-based and region-based ones [6]. The former extract specific image features like edges, corners [7], [11], scale invariant feature transform [12], [13], speeded up robust features [14], etc. Meanwhile, these algorithms are usually assisted by random sample consensus (RANSAC) [15] to eliminate mismatching pairs. A number of papers have addressed the image similarity modelling problems with region-based methods such as mutual information [16], grey correlation analysis [17], normalized cross correlation (NCC) [18] and alignment metric [19], which is applicable in pixel level image registration and similarity measurement. While region based registration algorithms require the pre-processing of image interpolation, it can be hardly applied to conduct image registration at sub-pixel level directly. In case of getting extra sub-pixel accuracy, two main techniques are adopted in previous works, consisting of interpolating the reference image (and /or the unmatched one) and optimizing the objective functions via surface fitting [10]. The accuracy of results would be affected by scaling factors (larger than 1) via interpolating and extreme value searching.

Another useful technique to align images is transforming the image from the spatial domain into frequency domain [20] via the method fast Fourier Transform (FFT) [21]. This approach is time consuming and requires large memory space, while Yousef et al. [22] uses single step discrete Fourier transform (SSDFT) with a huge reduction of computational complexity and memory requirement. It is effective to register images via phrase correlation methods [22], [23].

Won et al. [24] suggested minimizing the brightness change ratio and calculating the flow vector through Gauss-Seidel method, which minimizes the error for the ratio of brightness change between two images. Based on the previous works, this paper puts forward a fast image registration method by extended alignment metric at sub-pixel level via optimization problem solving to eliminate the flicker of blended faces in video. The maximizing alignment metric algorithm proposed by ref. [19] is used to convert the image registration into an optimization problem, which is extended to align images to be within a fraction of a pixel and locates optimal position directly in this work instead of search method or iteratively computing. This paper is organized as follows. In section 2, alignment metric is explained. In section 3, the framework of video face replacement and the proposed method for image registration are described. The behaviour of the algorithm is evaluated in section 4 and section 5 concludes the paper.

## 2. Alignment Metric

Alignment metric (AM) is a standard measurement to evaluate the registration level between two images. Since the aligned images may not have the same brightness value at same coordinates, alignment metric appears to handle different image modes according to ref. [19]. The greater the alignment metric value of two images is, the closer the similarity would be. Let $I_1(i, j)$ and $I_2(i, j)$ be the unregistered image and reference one respectively. The alignment metric, $AM[I_1(i,j), I_2(i,j)]$, is defined by ref. [19]

$$AM[I_1(i,j), I_2(i,j)] = \left( \frac{\overline{\sigma_{1,2}^2}}{\sigma_2^2} + \frac{\overline{\sigma_{2,1}^2}}{\sigma_1^2} \right)^{-1} . \quad (1)$$

Where $\sigma_k^2, k \in \{1,2\}$ is the intensity variance of the image, and $\overline{\sigma_{1,2}^2}$ is the interactive variance expectation of reference image relative to the unregistered one, with which the item $\overline{\sigma_{2,1}^2}$ can be illustrated by analogy conversely. Given the images, $I_1(i, j)$ and $I_2(i, j)$, $\overline{\sigma_{1,2}^2}$ is reached as

$$\overline{\sigma_{1,2}^2} = \sum_n p_1(n) \sigma_{1,2}^2(n), n = 0,1,2,\cdots,255 \quad (2)$$

Where $p_1(n)$ is the occurrence probability of intensity $n$, $\sigma_{1,2}^2(n)$ is the variance of reference image relative to the unregistered, and is the measurement of the intensity change of image $I_2(i, j)$ at the same coordinate in image $I_1(i, j)$. This process is given by Eq. (3),

$$\sigma_{1,2}^2(n) = \frac{1}{H_1(n)} \sum_{I_1(i,j)=n} \left( I_2(i,j) - \overline{E_{1,2}(n)} \right)^2 \quad (3)$$

Where item $H_1(n)$ is the histogram of image $I_1(i, j)$, and item $\overline{E_{1,2}(n)}$ is the average of $I_2(i, j)$ at the set of pixel that the intensity is $n$ in image $I_1(i, j)$. And item $\overline{E_{1,2}(n)}$ can be obtained by Eq. (4),

$$\overline{E_{1,2}(n)} = \frac{1}{H_1(n)} \sum_{I_1(i,j)=n} I_2(i,j) \quad (4)$$

The formula (3) measures the dispersion degree of the pixel sets in image $I_2(i, j)$ by the condition of the grey scale being n in image $I_1(i, j)$, vice versa; Equation (2) calculates the weighted results of Eq. (3) according to the probability of the pixel occurrence whose grey level is $n$ in image $I_1(i, j)$, and the average discrete degree of the pixel set is obtained with the same grey level in $I_2(i, j)$ corresponding to $I_1(i, j)$ in Eq. (1). The variance of the histogram of the other image is used as a standardized factor to get the interactive variance of the image normalized, and the similarity between the two images is measured by the inverse of the sum of the two standardized results. The larger the alignment metric is, the smaller the interactive variance is, which indicates the reference image and unmatched one get more similar.

On the whole, alignment metric is used to minimize the interleaving area of two images after overlapping. From another point of view, each grey level is most stable on the set of pixel positions corresponding to the same grey level compared with the other image. Mathematically, the interactive variance is the minimum and image registration task can be transformed into minimization problem.

## 3. Proposed Method

### 3.1 Framework of video face replacement

The framework of proposed video face replacement is shown in Fig.1, and it can be separated into 3 parts, face detection and landmarks location, video face alignment, and blending final frames. The techniques used and details are presented below.

*3.1.1 Face Locating and Tracking*: Firstly, the faces in source and target videos are detected by the algorithm proposed by Viola-Jones [25], which computes Haar-features, uses AdaBoost algorithm to construct strong classifiers, cascade them into a face classifier and gets high detection accuracy. The detected face area initializes the facial landmarks distribution and Supervised Descent Method (SDM) algorithm given in ref. [26] is exploited to fit the optimal location of facial feature points and head pose is obtained as well.

To track a face in a video sequence, the method of Struck algorithm [27] is used and facial features (e.g., eyes and nose) are set as the input spaces in a video clip. It is robust for different poses and expressions of the object and provides adaptive tracking between adjacent frames. From the second frame, it no longer needs to detect the location of the face unless the tracking face is missing.

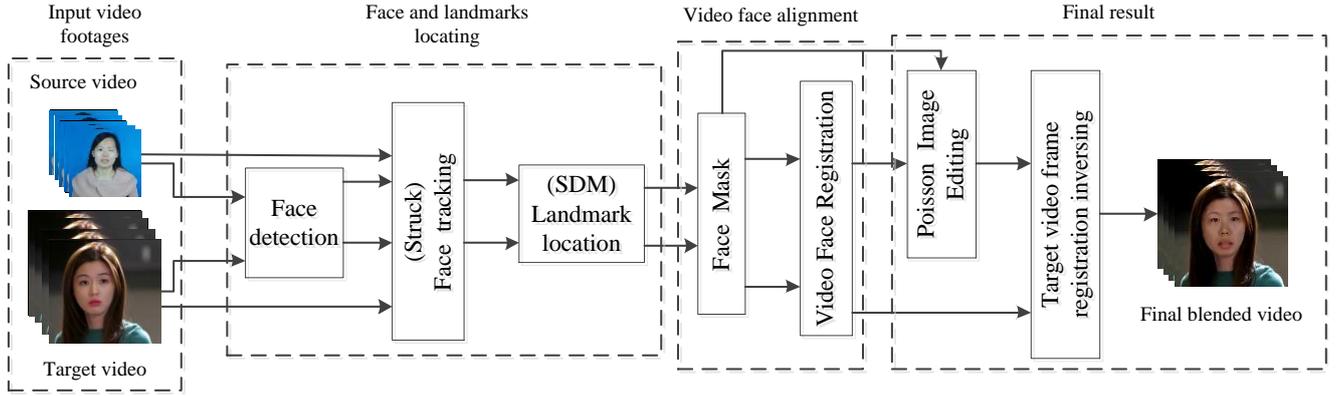

*Fig.1. Framework of proposed video face replacement*

*3.1.2 Alignment*: The alignment process proposed in this paper contains two parts, face alignment between adjacent frames along source and target video sequence and between source frames and target frames. And face registration approach along a video clip will be explained in sub-section 3.2 with experimental and theoretical analysis. In this stage, the video face is aligned to the first frame and the registered face in the resulted video sequence can be approximately seen as a straight pipe or a stream, in which the overlapping region of adjacent faces is the minimum. Due to the facial expressions and head movements, latter frame is aligned to the former within a limited error, which makes the registered video sequence piecewise smooth.

Despite the alignment along frame sequences, swapped result would not get natural footages without face match between source and target. In this stage, face landmarks are needed to calculate the affine parameters of the first frame, to resize and rotate the source frames ahead of registering the source face with the target. The method proposed by Torcida et al. [28] regarding feature point sets or shapes as input is used to get the parameters (scale, rotate matrix and translation vector) of the source face, which can get the head pose and face shape of source well matched with the target via coordinates transformation. Face mask is corresponding to the landmarks, blending face mask is obtained through the intersection of the mask from the source face landmarks and the one from the target in the first frame, and the face mask in the next frame is obtained by minimizing the alignment metric.

*3.1.3 Blending and alignment inversing*: Given registered source video frame and target video frame, the Poisson image editing algorithm proposed by ref. [29] can blend the two faces in a designated area (sub-region within the mask) and the target face is replaced by the source. In alignment process, the target video sequence is straightened out according to the face position, and after the face replacement, the straightened video frames need to be restored to the original position and size corresponding to the initial video sequence. The restoration is the inverse process of coordinate transformation in sub-section 3.1.2, with nothing changed except for the swapped faces.

### 3.2 Registration approach

Image registration to be discussed is a coarse-to-fine process; the best matching position is obtained by search method at integer pixel level and down sampling strategy is used to reach a high time efficiency, which is explained in sub-section 3.2.1. Accurate sub-pixel registration is calculated by minimizing extended alignment metric and solving equations derived from differential equations along x- and y- directions, which is described in sub-section 3.2.2.

*3.2.1 Coarse Pixel matching*: The similarity of the reference image and the unregistered can be evaluated by alignment metric as long as these two images have the same dimensions (same rows and columns). That is, the size of slide window is identical to that of the observed object and an alignment metric value, AM, could be reckoned within one scan, which denotes the similarity between the sub-image in the slide window and observed object. One course of pixel matching is the formulation of matrix composed of AMs to establish a match of the sub-image in slide window with the given image. Generally, the matrix is simply obtained by moving the slide window through the unregistered image step by step, which is time-consuming and calculates a dense matrix with a number of unnecessary figures. As is shown in Fig.2, the data of AM matrix are projected into an interval of [0, 1] obtained by the template right eye scanning the whole face. Obviously, much area is flat or gets a local maximum except small areas getting well matched, which indicates that the matrix can be set as a sparse one.

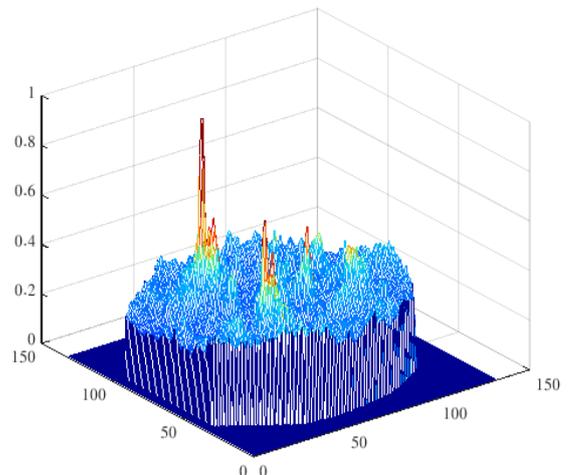

*Fig.2. Search Model of Alignment Metric of Right Eye Template*

In the process of calculating AM, a slide window goes through the unmatched image one pixel by one pixel. The image would be traversed repeatedly in which the time efficiency is low. In this section, according to the local continuity of images, down sampling approach is exploited for speeding up the maximum search process and creating a sparse matrix instead of a dense one. The search window moves in the candidate area of the image to be registered, which is initialized at the beginning and contains more pixels than the reference image. The down sampling method, by the way, directly operates on the initialized images and reconstructs the two images satisfying the precondition of keeping pixels from even rows and columns successively. An image pyramid is formulated by down sampling repeatedly and each layer is produced from the next layer in the same way while the original image is its bottom. The layers of image pyramid is limited by the size of the top template image whose minimum size is better than 8 according to ref. [30]. Each layer consists of two images, $I_1$ $(i, j)$ and $I_2$ $(i, j)$, a series of AM data would be calculated with the latter scanning the former by Eq. (1). Search region of a layer is constrained by the previous one observing the 3σ principle, which would dramatically reduce the computational complexity. And the position of maximum of AM is corresponding to coarse pixel matching coordinate. The AM peak value's 3σ area contains the maximum of AM in original image. And peak values are surrounded by flat area, which certainly explains the AM matrix is sparse and only a small part is most needed.

*3.2.2 Extended alignment metric*: Given two images $I_1$ $(i, j)$ and $I_2$ $(i, j)$ with the same dimension, alignment metric can easily measure their similarity by Eq. (1). And best match can be obtained in an image set with one image sized larger as explained in sub-section 3.2.1. After getting motion parameters between two images at integer pixel level, maximizing alignment metric method given by Eq. (1) is quite unable to deal with the registration accuracy at sub-pixel level, while the image grey level $n$ is critical to the AM calculation relating to Eq. (1)-(4) besides the sub-pixel value being unknown. Eq. (1) can be exploited to measure two images as long as both of them are constant images. While lateral shift is introduced, Eq. (1) can hardly deal with it. In this stage, extended alignment metric is explained to solve the registration problem and analysis method is used to calculate the root of proposed optimization problem. Suppose that unmatched image would not get well matched with the reference before performing a movement in less than one pixel. Similarly, reference image takes the opposite motion and registration task can be completed rapidly. Note that the difference of histogram is slight between the shifted image generated by interpolation and the original due to the image's local continuity, when an image is shifted with the motion parameters less than 1 pixel along x- and y-directions.

Suppose while image $I_1$ $(i, j)$ shifts a little step $\boldsymbol{d} = (x, y)^T, -1 < x, y < 1$, image $I_1$ $(i, j)$ and $I_2$ $(i, j)$ can get the peak value of AM, and the sub-pixel value is continuous between adjacent pixels. Image $I_2$ $(i, j)$ shifts $-\boldsymbol{d}$ and they can be registered similarly. Input shifted image $I_t(i+x, j+y)$ ($t$ = "1" denotes shifted image $I_1$ $(i, j)$ and $t$ ="2" $I_2$ $(i, j)$) and original image $I_s(i, j)$ ($s$ = "1" denotes original image $I_1$ $(i, j)$ and $s$ ="2" $I_2$ $(i, j)$, $s \neq t$ ) into Eq. (1) and AM gets the peak. It is a single step to calculate the sub-pixel motion instead of searching maximum step by step.

Lateral shift $\boldsymbol{d}$ is set to be an unknown or undetermined parameter to express the displacement of $I_t(i, j)$ from the most registered position with $I_s$ $(i, j)$ in the level below an integral pixel. Interactive variance expectation of images, $\sigma_{t,s}^2$ ($t$ and $s$ as a substitute for either "1" = "unmatched image" or "2" = "reference image"), is also the function of undetermined motion parameters. When unmatched image shifts, the reference image remains unchanged and each pixel appears to be a constant. And item $\sigma_{1,2}^2$ presents as a function of motion parameters. Conversely, the item $\sigma_{2,1}^2$ is the same. Eq. (1) can be written as a function of unknown motion parameters. As a result, sub-pixel registration is transformed into finding function extreme point and can be solved by use of analytical optimization method other than search method.

Generally, the search accuracy and efficiency of optimization performance both depend on the moving stride of search algorithms. In this work, analytic method is implemented. Suppose unregistered image $I_1$ $(i, j)$ differs from the reference $I_2$ $(i, j)$ by a drift offset $x$ and $y$ respectively along x- and y- directions, thus $x$ and $y$ can be computed as the following,

$$\arg\max_{x,y} AM[I_1, I_2], \quad x, y \in (-1,1). \quad (5)$$

When the alignment metric reaches its peak, the images line up exactly. Construct AM as the function of the unknown motion parameters, and let the partial derivative of the function in the x and y direction be equal to zero for equation set construction. Motion parameters are obtained by solving high order equation utilizing the method as is given by ref. [31].

It would be uneasy to get the derivative of Eq. (1). Note that cross variance CI is the inverse of AM.

$$AM[I_1, I_2] = \frac{1}{CI[I_1, I_2]} \quad (6)$$

Solving the maximum AM can be transformed into minimizing the cross variance problem, that is, minimize CI to formulate the equation set instead of getting maximum of AM. Thus Eq. (5) can be replaced by

$$\arg\min_{x,y} CI[I_1, I_2], \quad x, y \in (-1,1). \quad (7)$$

Cross variance has two components, both of which are normalized interactive variance expectations given by Eq. (2). Meanwhile, the normalization factors are the two images' variance which could eliminate the effect of illumination and quantized grey level.

Suppose intensities change uniformly among adjacent pixels and the intensity of shifted image can be substituted by the weighted mean of its near pixels. Since the intensity between two neighbouring pixels is well-distributed, bilinear interpolation is exploited for reconstructing the two images and is the bridge between motion parameters, $x$ and $y$, and the function CI. While calculates $\sigma_{1,2}^2(n)$ and $\sigma_{2,1}^2(n)$, one image is a scalar function, another constant image.

Since the pixel values of two-dimensional digital image are discrete, it is necessary to interpolate four regions (I, II, III, and IV) around each pixel. As is shown in Fig.3, a pixel is surrounded by four different regions with 8 neighbouring pixels differing from each other.

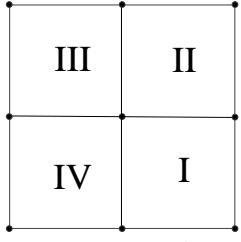

**Fig.3.** *Interpolation Regions of Each Pixel*

The components of the offset are different in four interpolation regions, respectively

$$\begin{cases} \text{I}: & x>0, y>0; \\ \text{II}: & x<0, y>0; \\ \text{III}: & x<0, y<0; \\ \text{IV}: & x>0, y<0. \end{cases} \quad (8)$$

In order to ensure the unified expression of the interpolation results, it is possible to set the two components of the offset with positive values $x, y$ ($x, y>0$). Movement of the scene in two images is relative. When one image is reconstructed sub-pixel in a region, another needs to be reconstructed in the corresponding opposite region. Take the pair, region I and III, as an example. While $\overline{\sigma_{1,2}^2}$ is constructed as the function of unknown motion parameter $d$ with interpolated scalar function $I_2(i, j)$ in region III and constant image $I_1(i, j)$, $\overline{\sigma_{2,1}^2}$ is formulated as the function of sub-pixel displacement with interpolated scalar function $I_1(i, j)$ in region I and constant image $I_2(i, j)$. Eq. (9) describes how a pixel reconstructs pixel with unknown parameters in region I of $I_1(i, j)$ shown in Fig. 4(a), and Eq. (10) in region III of $I_2(i, j)$ shown in Fig. 4(b).

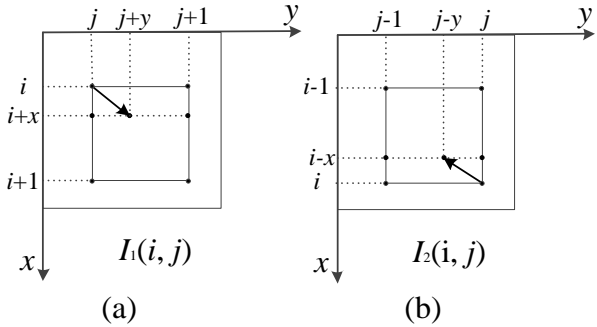

**Fig.4.** *Bilinear Interpolation*
**(a)** The interpolation of image $I_1(i, j)$ in region I;
**(b)** The interpolation of image $I_2(i, j)$ in region I

$$\begin{aligned} &I_1(i+x, j+y) \\ &= xy \times (I_1(i+1, j+1) - I_1(i, j+1) - I_1(i+1, j) + I_1(i, j)) \\ &+ x \times (I_1(i+1, j) - I_1(i, j)) + y \times (I_1(i, j+1) - I_1(i, j)) + I_1(i, j) \\ &= a_{ij} \times xy + b_{ij} \times x + c_{ij} \times y + d_{ij} \end{aligned} \quad (9)$$

$$\begin{aligned} &I_2(i-x, j-y) \\ &= xy \times (I_2(i-1, j-1) - I_2(i, j-1) - I_2(i-1, j) + I_2(i, j)) \\ &+ x \times (I_2(i-1, j) - I_2(i, j)) + y \times (I_2(i, j-1) - I_2(i, j)) + I_2(i, j) \\ &= e_{ij} \times xy + f_{ij} \times x + g_{ij} \times y + h_{ij} \end{aligned} \quad (10)$$

After interpolating the two images, $\overline{\sigma_{1,2}^2}$ and $\overline{\sigma_{2,1}^2}$ consist of $x$ and $y$, see (11) and (12), which can be replaced by two variable functions $f(x, y)$ and $g(x, y)$, respectively. And cross variance CI takes the following form,

$$CI[I_1, I_2] = \frac{\overline{\sigma_{1,2}^2}}{\sigma_2^2} + \frac{\overline{\sigma_{2,1}^2}}{\sigma_1^2} \quad (13)$$

$$= \frac{1}{MN\sigma_2^2} f(x, y) + \frac{1}{MN\sigma_1^2} g(x, y)$$

And functions $f(x, y)$ and $g(x, y)$ is given as follows,

$$\begin{cases} f(x, y) = \sum_n \sum_{i,j} \left( e'_{ij} xy + f'_{ij} x + g'_{ij} y + h'_{ij} \right)^2 \\ g(x, y) = \sum_n \sum_{i,j} \left( a'_{ij} xy + b'_{ij} x + c'_{ij} y + d'_{ij} \right)^2 \end{cases} \quad (14)$$

Parameters $a'_{ij}, b'_{ij}, \cdots, h'_{ij}$ are all constants in Eq. (14) and are composed of scalars $a_{ij}, b_{ij}, \cdots, h_{ij}$ in Eq. (9) and (10). Note that $CI[I_1, I_2]$ depends on unknown values $x$ and $y$, and the shift parameters $x$ and $y$ which satisfy optimized $CI[I_1, I_2]$ are obtained by letting

$$\begin{cases} \dfrac{\partial CI[I_1, I_2]}{\partial x} = \dfrac{1}{\sigma_2^2} f_x(x, y) + \dfrac{1}{\sigma_1^2} g_x(x, y) = 0 \\ \dfrac{\partial CI[I_1, I_2]}{\partial y} = \dfrac{1}{\sigma_2^2} f_y(x, y) + \dfrac{1}{\sigma_1^2} g_y(x, y) = 0 \end{cases} \quad (15)$$

Eq. (15) can be written as follows,

$$\begin{cases} \alpha_1 xy^2 + \alpha_2 xy + \alpha_3 y^2 + \alpha_4 y + \alpha_5 x + \alpha_6 = 0 \\ \beta_1 x^2 y + \beta_2 xy + \beta_3 x^2 + \beta_4 x + \beta_5 y + \beta_6 = 0 \end{cases} \quad (16)$$

And the coefficients of the unknown items $\alpha_k$ and $\beta_k (1 \le k \le 6)$ consist of coefficients $a'_{ij}, b'_{ij}, \cdots, h'_{ij}$, which can be obtained from original image $I_1(i, j)$ and $I_2(i, j)$. After

$$\overline{\sigma_{1,2}^2} = \frac{1}{MN} \sum_n \sum_{I_1(i,j)=n} \left( e_{ij} \times xy + f_{ij} \times x + g_{ij} \times y + h_{ij} - \frac{1}{H_1(n)} \sum_{I_1(k,l)=n} (e_{kl} \times xy + f_{kl} \times x + g_{kl} \times y + h_{kl}) \right)^2 \quad (11)$$

$$\overline{\sigma_{2,1}^2} = \frac{1}{M \times N} \sum_n \sum_{I_2(i,j)=n} \left( a_{ij} \times xy + b_{ij} \times x + c_{ij} \times y + d_{ij} - \frac{1}{H_2(n)} \sum_{I_2(k,l)=n} (a_{kl} \times xy + b_{kl} \times x + c_{kl} \times y + d_{kl}) \right)^2 \quad (12)$$

variable substitution in Eq. (16), a five-order equation with some certain preconditions is produced and can be solved with the method given by ref. [31]. There is only one undetermined variable left after variable substitution, maybe $x$ or $y$. For example, Eq. (16) can be written as

$$A_1 y^5 + A_2 y^4 + A_3 y^3 + A_4 y^2 + A_5 y + A_6 = 0 \qquad (17)$$
$$s.t. \alpha_1 y^2 + \alpha_2 y + \alpha_5 \neq 0$$

Items $A_k$ are composed of $\alpha_1, \alpha_2, \cdots, \alpha_6, \beta_1, \beta_2, \cdots, \beta_6$, and the horizontal shift along x- direction is given by Eq. (18).

$$x = -\frac{\alpha_3 y^2 + \alpha_4 y + \alpha_6}{\alpha_1 y^2 + \alpha_2 y + \alpha_5} \qquad (18)$$

Generally, only one valid root is obtained and maximizing alignment metric handles the case of more than one valid result. Translate the unregistered image according to motion parameters. Therefore, the process of sub-pixel registration is done on the misplaced image and the reference image.

## 4. Experimental Results

To evaluate the performance of the proposed approach, a series of experiments are conducted in two parts. Firstly, the framework suggested in this paper is exploited to replace video faces compared with two other works without face registration to eliminate the flicker or jitter of swapped faces in video sequence. In the second part, the registration method proposed is tested with a segmentation ground truth and the time efficiency and registration accuracy is analysed compared with previous works using a computer with an Intel Core i5, 2.50GHz processor and 6GB RAM via the development tool of Visual Studio 2013.

### 4.1 Effectiveness of Proposed Video Face Replacement Method

For no head motion videos, like some facial expression changed video in Max Planck Institute video face database [32], it is useful to verify the effectiveness of approaches in face registration. Table 1 shows the jitter variance of match points along video sequence, in which the actor's head does not move. The jitter variance denotes the stability of match points, and the variance is positively correlated to point jitter.

**Table 1** The jitter variance of match points

| algorithm | X direction | Y direction |
|---|---|---|
| Ref. [1] | 1.73 | 2.16 |
| Ref. [5] | 2.28 | 7.95 |
| Alignment metric | 1.23 | 1.43 |
| Extend alignment metric | 1.02 | 1.15 |

While the test is conducted on videos with actors' head moving, the reference points or ground truth does not exist. The ground truth of this experiment is obtained through manual marking, and each point is marked more than ten times. Let the average of points coordinates be the ground truth. There are more than 200 frames in the video clip, in which frames ranking from 51 to 150 are used. The Euclidean distance between ground truth and match points obtained from different methods can be calculated to measure the offset of observed point. And the degree of flicker is obviously shown below. Fig. 5 shows the stable flicker curve of the point offset of the proposed method compared with the methods given by ref. [1] and [5]. The offset in sub-graph (a) and (b) is much bigger than that in (c). Besides the values of the interval are greater than 2 in (a) and (b), and it approximately gets an interval of 1.5 in (c).

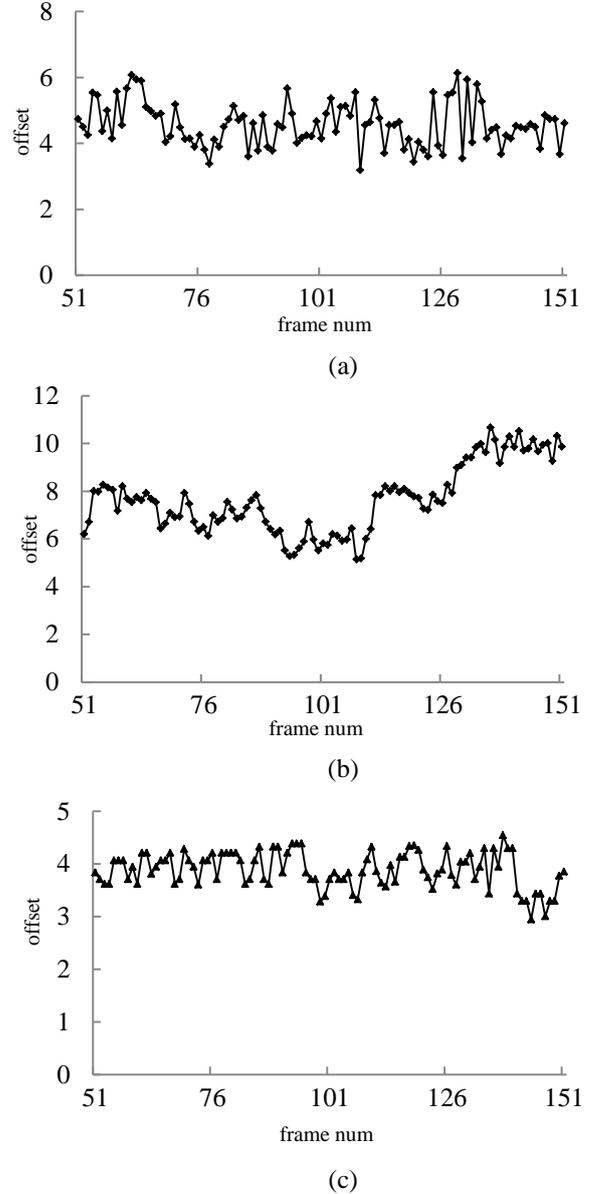

*Fig.5.* The point offset of different approaches
*(a)* The point offset of ref. [1], *(b)* The point offset of ref.[5], *(c)* The point offset of proposed method

### 4.2 The performance of sub-pixel registration

While registration method plays an important role in video face replacement to eliminate the flicker of swapped faces, it still does better in many applications like motion detection, super-resolution reconstructions, etc. The video frames and their ground truth are given bellow as sub-images (a) - (d) in Fig.6. According to Fig.7 (a), pixels in unmatched images (a) and (b) in Fig. 8 differ from those in reference image with the same coordinates. The translation vector of computational results between images (a) and (b) in Fig.6 is $(1.183645, 0.922107)^T$, and the motion vector of integer pixel ground truth is $(1, 1)^T$. While the ground truth may be not accurate, the calculation result is approximately consistent with it.

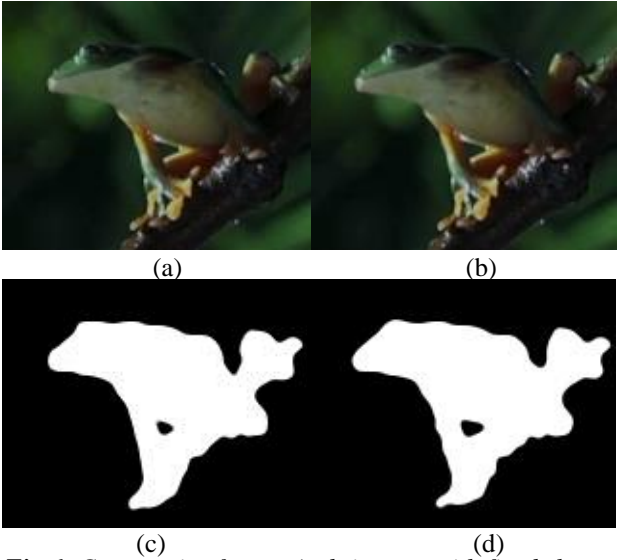

*Fig.6. Consecutive frames (sub-images with fixed shot and the corresponding ground truth)*
(*a*) Previous frame, (*b*) the next frame, (*c*) moving object in reference image, (*d*) moving object in unmatched image

The proposed method is applicable to image registration as is shown in Fig. 7(b). At the beginning, a coordinate set is obtained with each point randomly picked in the matching area of two consecutive frames in a video clip given by [33] (the previous one as reference and the next frame unmatched) within a small size of $80 \times 80$, and the corresponding intensities are picked simultaneously. Due to the motion of observed objects, brightness would be different at the same position in adjacent frames shown in Fig. 7 (a). Hence, the translation vector can be obtained from the difference between them. Fig.7 (b) shows pixels are well matched after the implementation of the proposed method.

The computational complexity of this approach is $O\left(n/4^l + \lambda(n) + \alpha n\right)$, where $n$ is the number of pixels in the image to be registered and $l$ is the times of down sampling. The first item of computational complexity means the time consummation of calculating AMs at last down sampling, the second item refers to time cost to form the sparse matrix, and the third spend of aligning two images with sub-pixel accuracy. It is acceptable to merge the first and the third items, and computational complexity of this approach can be written as $O(\lambda(n)+n)$, which is a great reduction over that of usual approaches. Compared with the method by Jackson et al. [11], Yu et al. [12], the proposed method is conducted in different dimensions at the image set with high resolution. Owing to the enormous memory required, the approach proposed by Yu et al. cannot handle an image larger than $1000 \times 1000$.

**Table 2** Execution time (in seconds) of different approaches

| Dimensions | Jackson et al. | Yu et.al. | Proposed |
|---|---|---|---|
| $100 \times 100$ | 0.044 | 0.14 | 0.044 |
| $200 \times 200$ | 0.1 | 0.28 | 0.069 |
| $400 \times 400$ | 0.244 | 0.692 | 0.186 |
| $800 \times 800$ | 0.808 | 2.7 | 0.561 |
| $1000 \times 1000$ | 1.68 | 3.776 | 0.956 |
| $2000 \times 2000$ | 5.716 | - | 3.862 |
| $3000 \times 3000$ | 6.032 | - | 4.937 |

The time efficiency comparison between proposed method and the approaches in the reference [11] and [12] is displaced in table 2, and each experiment can get a sub-pixel shift. In addition, the method normalized cross correlation (NCC) given by ref. [18] would calculate the motion parameters at integer accuracy, and is also time consuming because convolution process is involved with a high computational complexity.

Since the time efficiency is high, the sub-pixel image registration would be evaluated by accuracy. In this part, experimental samples are generated by shifting and down sampling a single image in test set, and coherent lateral shifts are set as their ground truth. Sub-pixel shift can be estimated by the method given by Song et al. [18], Yu et al. [12] and the proposed. Accuracy (or max errors) of different approaches can be assessed by comparing the calculation results in each image set against the ground truth at hand. An unregistered image can be reached using coordinate transformation when the image shifts with integer pixel and an image would be reconstructed by interpolation while the shift parameters are with decimal fraction. Therefore, the max errors would be affected by interpolation when omitting decimal fractions smaller than 0.5 and counting all others, including 0.5, as 1. Fig. 8 shows the max errors of different methods in the literature affected by interpolating process as

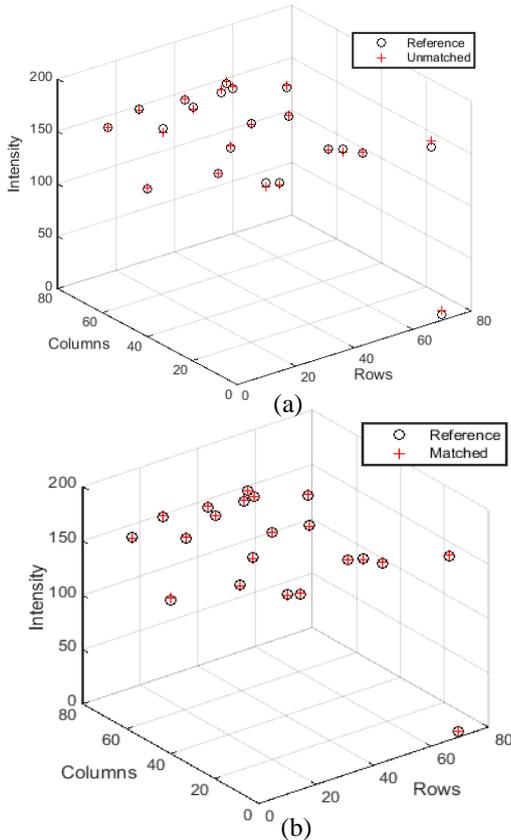

*Fig.7. Data before and after registration (pixel location is randomly picked)*
(*a*) Unmatched data of two images, (*b*) Data of registered images an idea given in ref. [34]. Calculation results get closer to the ground truth when image motion with a decimal fraction near 0.5 and max errors get closer to 0 as can be seen in fig. 8. It can be got that the max errors of proposed approach are smaller and absolute max error is 0.03 approximately.

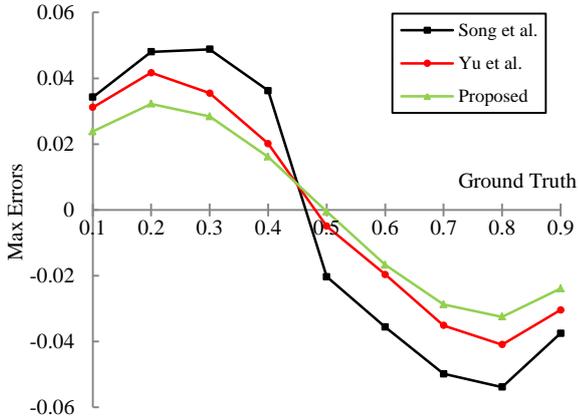

*Fig.8. Registration accuracy of proposed approach*

Experimental data in fig. 8 are obtained with an image set sized $396 \times 396$, and the max errors would be much smaller as simulation test is conducted on larger images. In addition, max errors of horizontal and vertical directions can converge to 0.01, which can be explained by a bias originated from the interpolation of the discrete image. Images are shifted at the motion vector with a fraction of 0.5 to get rid of that effect; therefore, an estimator in ref. [34] is exploited to quantitatively assess the accuracy of different techniques. Meanwhile, the performance of registration techniques is discussed with different Gaussian white noise ranging from 0 to 3. The estimation efficiency of different registration techniques is shown in table 3. According to Uss et al. [34], registration algorithm is more accurate if the estimator efficiency is close to 1. Additionally, the method SIFT given by Yu et al. [12] and proposed is robust to Gaussian white noise.

**Table 3** Estimator efficiency for different registration techniques with Gaussian white noise

| Sigma of Gaussian noise | Song et.al. | Yu et.al. | Proposed |
| --- | --- | --- | --- |
| 0 | 0.986719 | 0.901291 | 0.998631 |
| 1 | 0.963451 | 0.896173 | 0.993742 |
| 2 | 0.956285 | 0.894586 | 0.998956 |
| 3 | 0.938716 | 0.882697 | 0.998966 |

## 5. Conclusion

This paper presents a framework of video face replacement for solving the flicker of swapped faces in video sequence via extended alignment metric to register faces. The framework contains two steps of alignment and an inversing process of alignment: one is to get frames along a video sequence well matched and another is to get the source face well aligned with the target. The inversing step is to restore the position of the frames in target video being registered along frame sequence. And a fast sub-pixel registration algorithm based on image similarity measurement is proposed, which is divided into integer pixel registration and sub-pixel registration. Integer pixel registration is based on maximizing the AM via down sampling strategy to speed up the search process; sub-pixel registration is a single-step process, using bilinear interpolation and unknown accurate coordinate to reconstruct the registered image, transforming the image registration problem into an optimization problem and solving the problem via calculating resolutions of high order equations. The algorithm overcomes the iterative solution of sub-pixel displacement in the process of time cost, reduces the error of image registration, and presents a new method of calculating sub-pixel displacement, in real-time with high accuracy on the basis of extended alignment metric and analytic method.